\documentclass[10pt,twocolumn,letterpaper]{article}

\usepackage[pagenumbers]{cvpr} 

\usepackage{graphicx}
\usepackage{amsmath}
\usepackage{amssymb}
\usepackage{booktabs}
\usepackage{footmisc}
\usepackage{dblfnote}

\newcommand{\mycomment}[1]{}

\usepackage[pagebackref,breaklinks,colorlinks]{hyperref}

\usepackage[capitalize]{cleveref}
\crefname{section}{Sec.}{Secs.}
\Crefname{section}{Section}{Sections}
\Crefname{table}{Table}{Tables}
\crefname{table}{Tab.}{Tabs.}

\def\cvprPaperID{5382} 
\def\confName{CVPR}
\def\confYear{2023}

\begin{document}

\title{QuadFormer: Quadruple Transformer for Unsupervised Domain Adaptation in Power Line Segmentation of Aerial Images}
\author{\textbf{Pratyaksh Prabhav Rao}$^{1,2\ast}$, \textbf{Feng Qiao}$^{2\ast}$, \textbf{Weide Zhang}$^{2}$, \textbf{Yiliang Xu}$^{2}$, \textbf{Yong Deng}$^{2}$, \\
\textbf{Guangbin Wu}$^{2}$, 
\textbf{Qiang Zhang}$^{2}$ \\
$^{1}$New York University $^{2}$Autel Robotics\\
}
\maketitle

\begin{abstract}
Accurate segmentation of power lines in aerial images is essential to ensure the flight safety of aerial vehicles. Acquiring high-quality ground truth annotations for training a deep learning model is a laborious process. Therefore, developing algorithms that can leverage knowledge from labelled synthetic data to unlabelled real images is highly demanded. This process is studied in Unsupervised domain adaptation (UDA). Recent approaches to self-training have achieved remarkable performance in UDA for semantic segmentation, which trains a model with pseudo labels on the target domain. However, the pseudo labels are noisy due to a discrepancy in the two data distributions. We identify that context dependency is important for bridging this domain gap. Motivated by this, we propose QuadFormer, a novel framework designed for domain adaptive semantic segmentation. The hierarchical quadruple transformer combines cross-attention and self-attention mechanisms to adapt transferable context. Based on cross-attentive and self-attentive feature representations, we introduce a pseudo label correction scheme to online denoise the pseudo labels and reduce the domain gap. Additionally, we present two datasets - ARPLSyn and ARPLReal to further advance research in unsupervised domain adaptive powerline segmentation. Finally, experimental results indicate that our method achieves state-of-the-art performance for the domain adaptive power line segmentation on ARPLSyn$\rightarrow${TTTPLA} and ARPLSyn$\rightarrow${ARPLReal}.   

\end{abstract}

\section{Introduction}
\label{sec:intro}

\def\thefootnote{$\ast$}\footnotetext{These authors contributed equally to this work.}

Unmanned Aerial Vehicles (UAVs) are being used in a plethora of different applications such as photography, commercial, military, agricultural, and exploration. One of the biggest challenges during the UAV flight is their inability to detect widespread power lines (PLs). Collision with PLs would not only damage the UAV but can adversely affect power grids. Hence, it is imperative that the UAVs are able to accurately detect PLs during flight. PL segmentation is challenging for two reasons. First, the geometric structure of PLs is very thin and spans only a tiny portion of the image. Therefore, the segmentation is vulnerable to being fragmented. Second, there is a lack of high-quality open-source datasets dedicated to PL segmentation. It is expensive and cumbersome to collect vast amounts of labelled data. One way to mitigate this issue is to transfer knowledge from models trained on synthetic datasets with freely-available ground truth annotations (source domain) to the unlabelled real images (target domain). However, while transferring knowledge from one domain to another, deep learning models suffer from poor generalization performance due to a domain shift (varied lighting conditions, color distribution, backgrounds, etc). 


\begin{figure}[t]
  \begin{center}
    \includegraphics[width=0.475\textwidth, trim={0 0 0 3.0cm},clip]{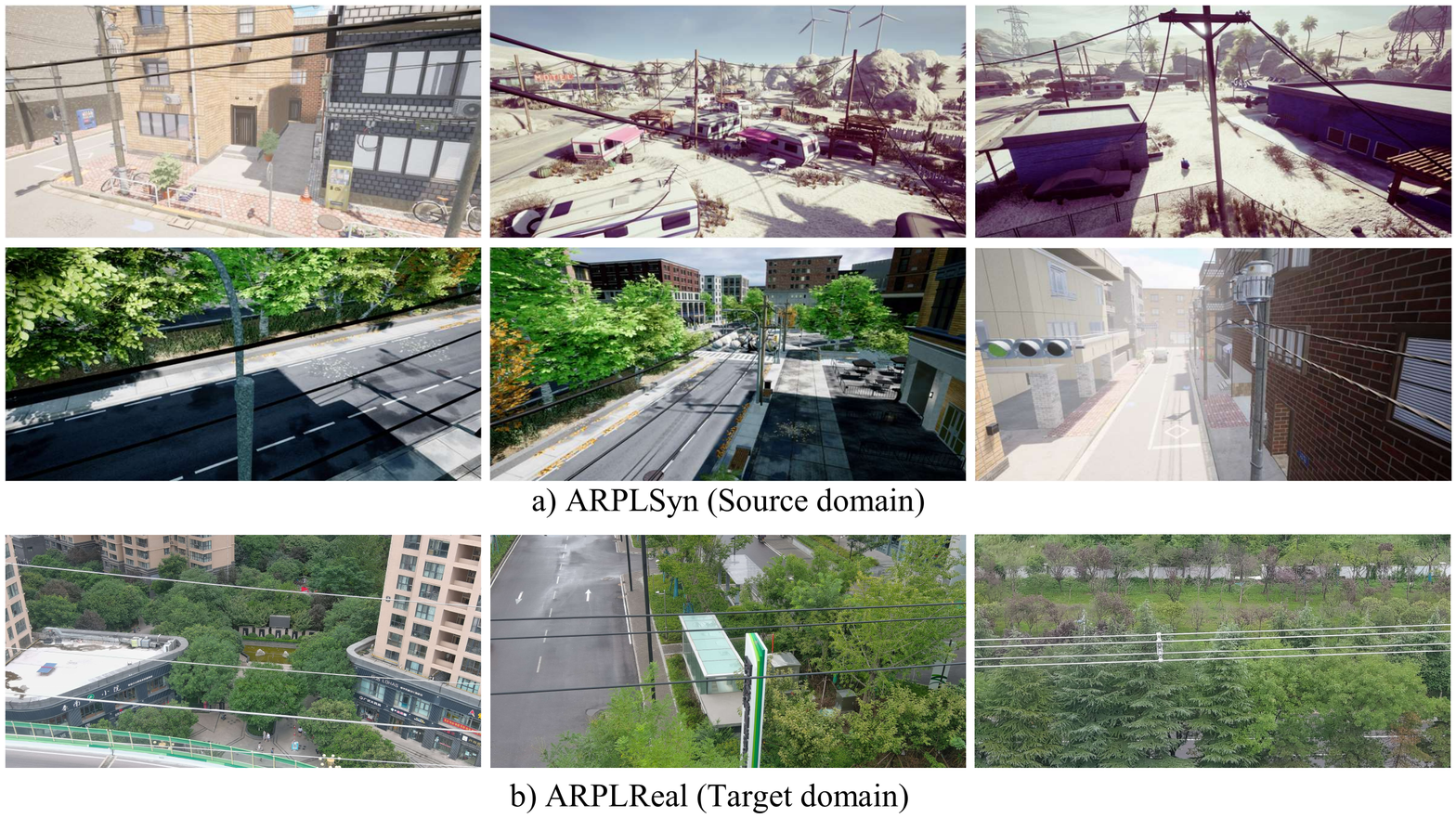}
    \vspace{-10mm}
    \caption{Examples from the a) ARPLSyn datset and b) ARPLReal dataset. The figure illustrates the diversity in both datsets, covering complex backgrounds, lighting conditions, zoom levels, and view angles.}
    \label{datasetexamples}
    \label{fig:HD_metric}
  \end{center}
  \vspace{-5mm}
\end{figure}

To address the domain gap problem, UDA techniques aim to align distributions of the labelled source domain and unlabelled target domain \cite{chang2019all, hong2018conditional, saito2018maximum, tsai2018learning, wan2020bringing}. Recently, self-training has emerged as a competitive approach in the UDA task \cite{li2019bidirectional, zhang2019category, zou2018unsupervised, zou2019confidence}. It involves iteratively generating pseudo labels on the target domain by a model trained on a labelled source domain and then utilizing these pseudo labels to retrain the network. However, due to the domain gap, pseudo labels are noisy. Thus, the network fails to learn reliable knowledge in the target distribution. Moreover, these methods fail to leverage explicit contextual information across different domains, which is essential for scene understanding \cite{zhang2018context, zhao2018psanet} and aligning different distributions \cite{yang2021context, xu2021cdtrans, wang2022domain}. While the appearance between the domains is different, the semantic context for PLs is similar. Therefore, cross-domain information is essential for generating robust representations for distinguishing PLs from their background. Motivated by this, we propose a novel UDA framework named \textit{QuadFormer}. The key idea of QuadFormer is to leverage cross-domain context dependencies to online denoise the pseudo labels. The hierarchical quadruple transformer combines cross-attention and self-attention for learning robust feature representations. The pseudo labels are corrected on the fly as the training progresses. While the self-attentive features provide rich contextual information to capture very thin PLs and enforce a global consistency between pixels, the cross-attentive features provide semantic context across different domains. Additionally, to further research in power line segmentation, we present two datasets - \textit{ARPLSyn} (synthetic) and \textit{ARPLReal} (real) with high-quality segmentation annotations (refer Figure.~\ref{datasetexamples}). The datasets cover a wide range of complex scenes, backgrounds, lighting conditions, and zoom levels.

In summary, our contributions are as follows - 

\begin{enumerate}
    \item We propose a hierarchical quadruple transformer that leverages self-attentive and cross-attentive features to adapt transferable context between the labelled source domain and unlabelled target domain.`
    \item Based on cross-attentive and self-attentive feature representations, we implement an online pseudo label correction mechanism to bridge the domain gap.
    \item We present two public datasets, ARPLSyn (synthetic) and ARPLReal (real) to promote research in unsupervised domain adaptive PL segmentation. 
    \item Extensive experiments indicate that our method achieves state-of-the-art performance for the domain adaptive power line segmentation on ARPLSyn$\rightarrow${TTPLA} (46.32 IoU) and ARPLSyn$\rightarrow${ARPLReal} (44.25 IoU).
\end{enumerate}




\section{Related Work}
\textbf{Unsupervised Domain Adaptation (UDA)}
UDA methods are classified into adversarial training and self-training. In adversarial training, the goal is to align the distribution of the source and target domain based on style-transferred inputs \cite{gong2021dlow, hoffman2018cycada}, network features \cite{hoffman2016fcns, tsai2018learning}, or outputs \cite{vu2019advent}. However, global alignment of the source and target distribution cannot guarantee optimal performance on the target domain. In self-training, the network is trained by computing pseudo-labels \cite{lee2013pseudo} for the target domain. Many methods pre-compute the pseudo labels during an offline warm-up stage and re-train the model \cite{yang2020fda, zou2018unsupervised, zou2019confidence}. On the other hand, pseudo labels can be computed during the training process. To ensure the selected samples are balanced, adaptive thresholds are utilized for semantic classes \cite{araslanov2021self, subhani2020learning}. In order to ensure that the training process is regularized, pseudo-label prototypes \cite{zhang2021prototypical}, or consistency regularization \cite{sohn2020fixmatch, tarvainen2017mean}, or domain mix-up are implemented \cite{tranheden2021dacs, zhou2022context}. The majority of past work utilizes DeepLabV2 \cite{chen2017deeplab} with ResNet-101 \cite{he2016deep} as the backbone. More recently, transformer-based architecture has proven to be powerful for UDA \cite{hoyer2022daformer, hoyer2022hrda}.  

\textbf{Context-Aware Embedding} Context plays an important role in scene understanding \cite{mottaghi2014role}. In order to capture the semantic context of scenes, Zhang \textit{et al.} \cite{zhang2018context} propose a context encoding module. Lin \textit{et al.} \cite{lin2018multi} utilize multi-scale context embedding for semantic segmentation.  \cite{yang2021context} adapt context dependencies from both spatial and channel views through self-attention and cross-attention mechanisms. Another line of work leverage cross-attention mechanisms via vision transformers for both feature learning and domain alignment \cite{xu2021cdtrans, wang2022domain}. We draw inspiration from BCAT \cite{wang2022domain} to generate mix-up features through the transformer-based cross-attention mechanism. Different from BCAT, we design a novel hierarchically structured Quadruple Transformer for semantic segmentation to adapt context dependencies across the source and target domain.

\textbf{PL Segmentation} Existing work on PL segmentation focus on traditional computer vision techniques \cite{golightly2005visual}, \cite{candamo2009detection}. Traditional methods rely on handcrafted features and predefined hyper-parameters to generate accurate results. However, this process is very challenging and cannot generalize to different aerial conditions. More recently, deep learning based methods are implemented \cite{madaan2017wire, zhao2019region} for PL segmentation. While the results are promising, these methods require large amounts of labelled ground truth annotations for training powerful models. In this work, we focus on the unsupervised domain adaptive PL segmentation task which does not require ground truth annotations for the target domain.

\textbf{Dataset} There have been several published and unpublished datasets on PLs. ImageNet~\cite{russakovsky2015imagenet} contains 1290 images with ground truth annotations on PLs, most of which are not captured by UAVs. In ~\cite{https://doi.org/10.17632/n6wrv4ry6v.8}, which is referred to as PLID, 2000 images containing PLs are presented. Zhang \textit{et al.} ~\cite{zhang2019detecting} present two datasets, covering different scenarios in urban and mountain settings. The wire detection dataset (WDD)~\cite{madaan2017wire} superimposes synthetic wires on 67K aerial images. TTPLAA~\cite{abdelfattah2020ttpla} is a publicly available PL aerial dataset containing 1100 high-resolution images and ground truth annotations. As far as the authors know, ARPLSyn is the first synthetic dataset for the PL segmentation task made publicly available.

\label{sec:related work}

\begin{figure}[t]
  \begin{center}
    \includegraphics[width=0.49\textwidth, trim={0 0 0 3cm},clip]{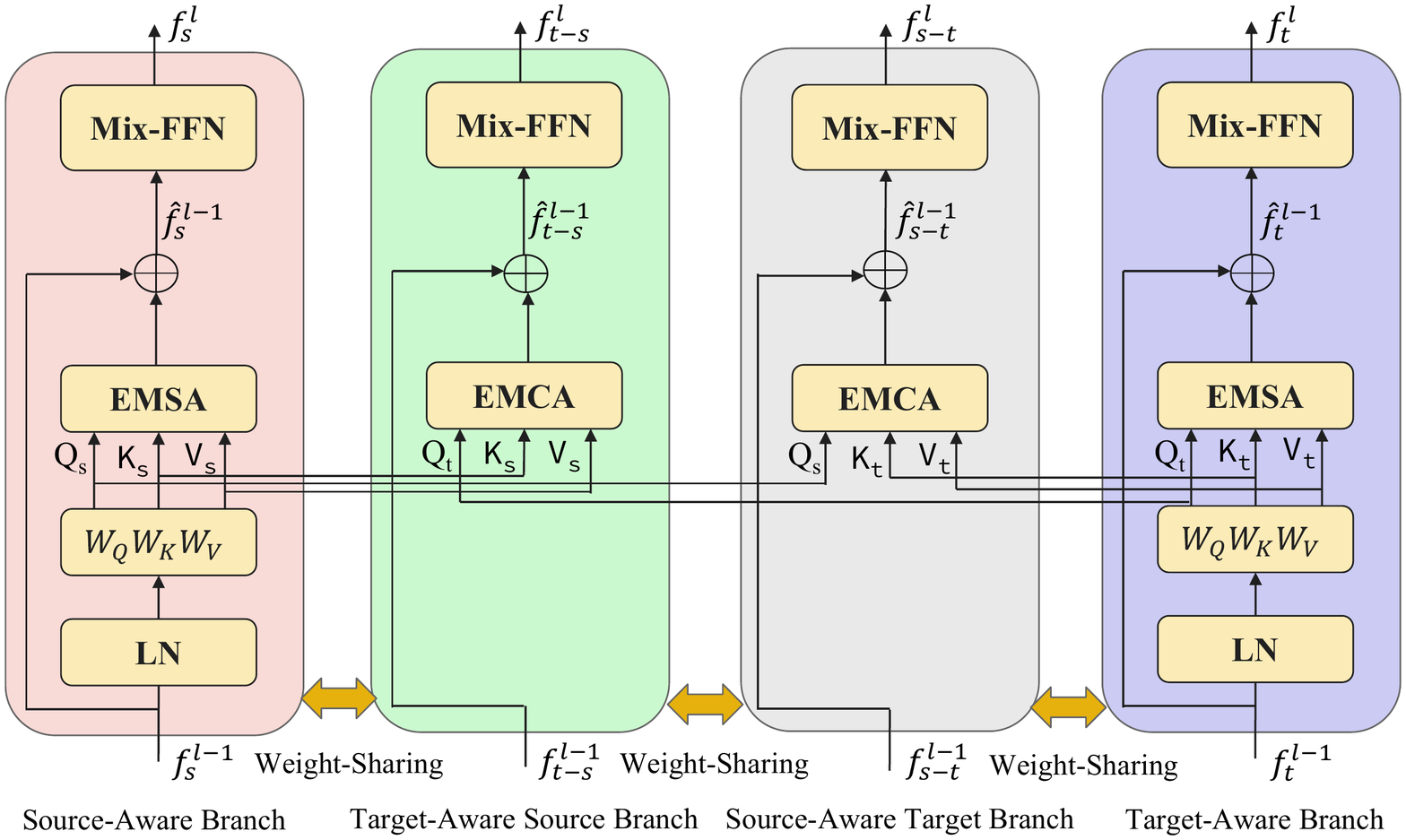}
    \vspace{-10mm}
    \caption{A quadruple transformer block for generating self-attentive and cross-attentive features. $W_{Q}, W_{K}, W_{V}$ are the transformation parameters to compute the queries, keys, and values, respectively.'LN' denotes layer normalization. 'EMSA' and 'EMCA' denote the efficient multi-head self-attention and efficient multi-head cross attention, respectively.}
    \label{uadruple_transforme}
  \end{center}
  \vspace{-6mm}
\end{figure} 

\begin{figure*}[t]
    \centering {\includegraphics[width=0.90\textwidth, trim={0 1.20cm 0 3cm},clip]{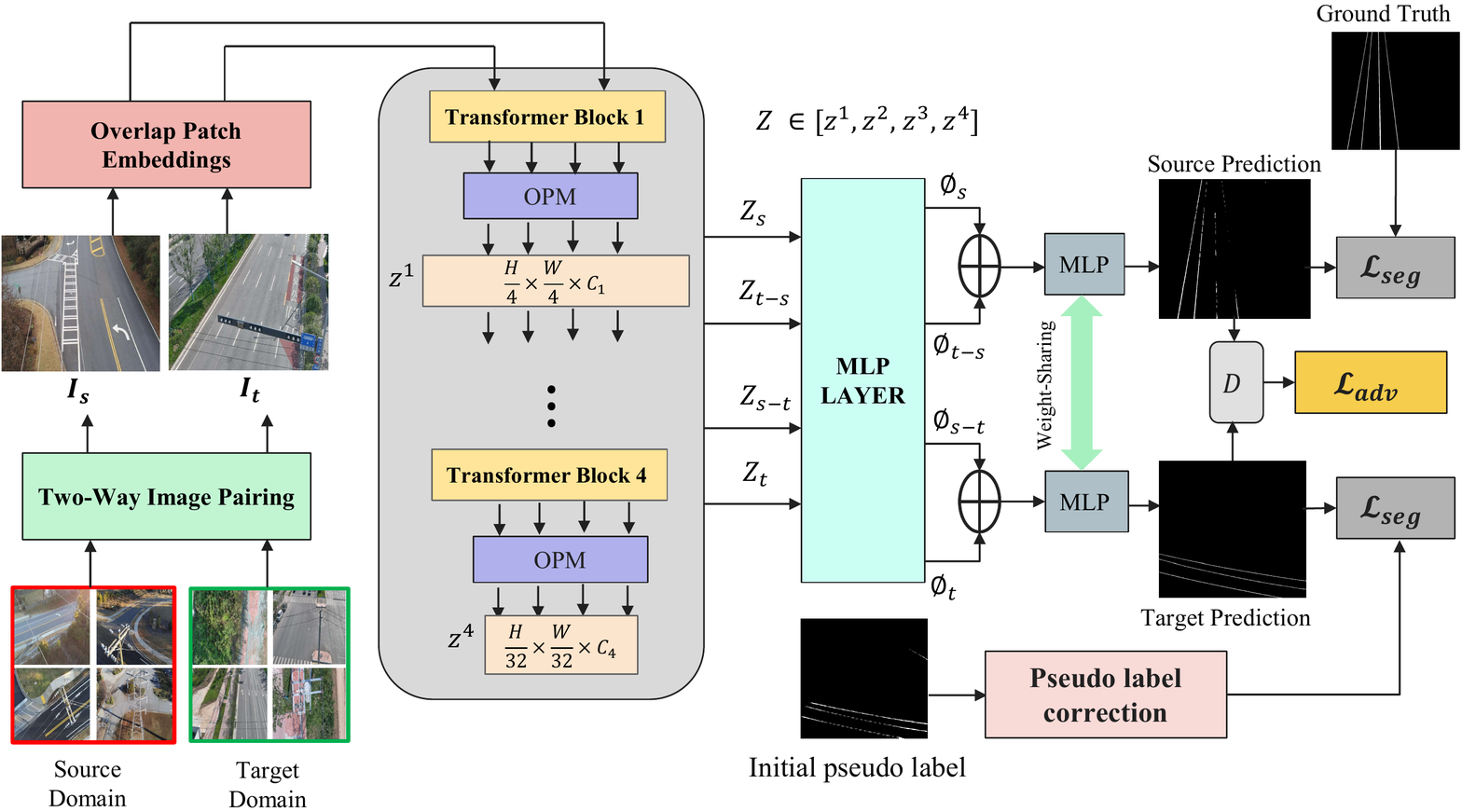}} \quad
    \vspace{-7mm}
    \caption{The proposed QuadFormer framework. Given an image pair (one from source domain and one from target domain), images are divided into patches of size  $4\times4$. These patches are fed to the hierarchical transformer encoder to obtain features at multiple scale, i.e $\{1/4, 1/8, 1/16, 1/32\}$ of the original resolution of the images. The hierarchical transformer encoder outputs multi-scale self-attentive and cross-attentive features which are then passed to the ALL-MLP decoder to generate four features vectors $\{\phi_{s}, \phi_{t-s}, \phi_{s-t}, \phi_{t} \}$. These features are utilized to obtain augmented representations to predict segmentation masks for the source and target domain. The pre-computed pseudo labels are corrected as the training progresses.}
    \label{workflow}
\vspace{-10mm}
\end{figure*}

\section{PL Dataset}
We present two PLs datasets with high-quality segmentation annotations. The ARPLSyn is collected by utilizing the Unreal Engine 5. High-resolution images and ground truth annotations are collected through the AirSim~\cite{shah2018airsim} plugin developed by Microsoft. Through the AirSim interface, we are able to obtain synchronized RGB stereo images, and segmentation labels. To ensure diversity, we set up 4 environments, covering a wide range of scenarios, view angles, light conditions, and zoom levels. Each environment has at least one sequence of images and their corresponding PL segmentation annotations. These images are then sampled once every 2 frames and manually inspected for the possibility of poor-quality images. The ARPLSyn dataset contains 7000 images and ground truth annotations of resolution $3840 \times 2160$.

The ARPLReal dataset is collected using a UAV. The onboard cameras provide high-resolution images ($3840 \times 2160$). To avoid duplicate images, we sample every other frame from the captured video. The dataset covers complex backgrounds, zooming angles, weather and lighting conditions. We manually inspect the dataset and discard unclear images. The ARPLReal dataset contains 3800 training and 200 validation images with ground truth annotations.


\section{Methodology}
\label{sec:methodology}

In this section, we start by summarizing the key idea of our approach. We then detail the proposed cross-domain transformer encoder and decoder for adapting contextual information between the source and target domain. Based on cross-attentive feature representations, we describe our pseudo label correction mechanism to online denoise the soft pseudo labels.

\subsection{Overview} 
For a UDA problem, we are given a source dataset $X_{s} = \{x_{s}\}_{j=1}^{n_{s}}$ with ground truth labels $Y_{s} = \{y_{s}\}_{j=1}^{n_{s}}$, and unlabelled target dataset $X_{t} = \{x_{t}\}_{j=1}^{n_{t}}$. The two distributions suffer from a domain shift. The goal of UDA is to train a segmentation model that can provide accurate predictions for $X_{t}$. Our training objective can be divided into two stages. First, motivated by the recent success of self-training\cite{li2019bidirectional}, we generate pseudo labels $Y^{st}_{t}$ for  $X_{t}$. Then, during the domain adaptation stage, the segmentation model is retrained with the labelled source dataset, target images, and pseudo labels. Figure.~\ref{workflow} illustrates the proposed UDA framework. The QuadFormer consists of a cross-domain transformer encoder (Section~\ref{cross_domain_encoder}) for generating self-attentive and cross-attentive multi-scale features, and a cross-domain decoder (Section~\ref{cross_domain_decoder}) for predicting segmentation masks for the source and target domain. Additionally, we implement a pseudo label correction mechanism to online denoise the pseudo labels.


\subsection{Cross-Domain Transformer Encoder}\label{cross_domain_encoder}
Mix Transformers (MiT) \cite{xie2021segformer} have achieved remarkable performance on semantic segmentation. One of the most important structures in MiT is the hierarchically structured transformer encoder which outputs multiscale features. Inspired by this, the QuadFormer method combines two self-attention and two cross-attention modules to design the four-branch transformer encoder. Given an image pair (one from the source domain and one from the target domain), images are divided into patches of size 4 × 4. The image patches are transformed into three vectors - queries $Q$, keys $K$, and values $V$. A sequence reduction process \cite{wang2021pyramid} is implemented to cope with high feature resolution and reduce the computational complexity of the attention computation. The self-attention is estimated as - 

\begin{gather}
    Attn_{self}(Q_{s}, K_{s}, V_{s}) = \textrm{Softmax}(\dfrac{Q_{s}K_{s}^{T}}{\sqrt{d_{head}}})V_{s}
\end{gather}

\begin{gather}
    Attn_{self}(Q_{t}, K_{t}, V_{t}) = \textrm{Softmax}(\dfrac{Q_{t}K_{t}^{T}}{\sqrt{d_{head}}})V_{t}
\end{gather}

\noindent where $d_{head}$ indicates the vector dimension, $Q_{s}$, $K_{s}$, $V_{s}$ are queries, keys, and values from the patches of image $I_{s}$, and $Q_{t}$, $K_{t}$, $V_{t}$ are queries, keys, and values from the patches of image $I_{t}$. The cross-attention operation is derived from the self-attention operation. We leverage this module to generate mix-up features and is formulated as - 

\begin{gather}
    Attn_{cross}(Q_{s}, K_{t}, V_{t}) = \textrm{Softmax}(\dfrac{Q_{s}K^{T}_{t}}{\sqrt{d_{head}}})V_{t}
\end{gather}

\begin{gather}
    Attn_{cross}(Q_{t}, K_{s}, V_{s}) = \textrm{Softmax}(\dfrac{Q_{t}K^{T}_{s}}{\sqrt{d_{head}}})V_{s}
\end{gather}

\vspace{-3mm}
\noindent To obtain hierarchical feature maps $F_{i} \in 	\mathbb{R}^{{H}/{2^{i+1}} \times {W}/{2^{i+1}} \times C_{i}}$, we perform patch merging\cite{xie2021segformer}. Additionally, since positional encoding is not necessary for the semantic segmentation task, we implement the Mix Feed-Forward Network (Mix-FFN) to leak local information \cite{islam2020much}. Mix-FFN integrates a $3\times3$ convolution and a Multi-Layer Perceptron (MLP) into a feed-forward network (FFN). Figure.~\ref{uadruple_transforme} illustrates the proposed quadruple transformer block and consists of four branches - a) source-aware branch, b) target-aware source branch, c) source-aware target branch, and d) target-aware branch. The self-attention module and Mix-FFN extract source-aware features $F_{s}$ and target-aware features $F_{t}$. Furthermore, with the cross-attention module and Mix-FFN, the cross-domain branches produce target-aware source features $F_{t-s}$ and source-aware target features $F_{s-t}$. The transformer block can be formulated as follows - 

\mycomment{
\begin{gather*}
    \hat{f}_{s}^{l} = \textrm{EMSA}(\textrm{LN}(f_{s}^{l-1})) + f_{s}^{l-1} 
\end{gather*}

\begin{gather*}
    f_{s}^{l} = \textrm{MLP}(\textrm{GELU}(\textrm{Conv}_{3 \times 3}(\textrm{MLP}(\hat{f}_{s}^{l})))) + \hat{f}_{s}^{l}
\end{gather*}

\begin{gather*}
    \hat{f}_{t}^{l} = \textrm{EMSA}(\textrm{LN}(f_{t}^{l-1})) + f_{t}^{l-1} 
\end{gather*}

\begin{gather*}
    f_{t}^{l} = \textrm{MLP}(\textrm{GELU}(\textrm{Conv}_{3 \times 3}(\textrm{MLP}(\hat{f}_{t}^{l})))) + \hat{f}_{t}^{l}
\end{gather*}

\begin{gather*}
    \hat{f}_{t-s}^{l} = \textrm{EMCA}(\textrm{LN}(f_{t}^{l-1}), \textrm{LN}(f_{s}^{l-1})) + f_{t-s}^{l-1} 
\end{gather*}

\begin{gather*}
    f_{t-s}^{l} = \textrm{MLP}(\textrm{GELU}(\textrm{Conv}_{3 \times 3}(\textrm{MLP}(\hat{f}_{t-s}^{l})))) + \hat{f}_{t-s}^{l}
\end{gather*}

\begin{gather*}
    \hat{f}_{s-t}^{l} = \textrm{EMCA}(\textrm{LN}(f_{s}^{l-1}), \textrm{LN}(f_{t}^{l-1})) + f_{s-t}^{l-1} 
\end{gather*}

\begin{gather*}
    f_{s-t}^{l} = \textrm{MLP}(\textrm{GELU}(\textrm{Conv}_{3 \times 3}(\textrm{MLP}(\hat{f}_{s-t}^{l})))) + \hat{f}_{s-t}^{l}
\end{gather*}
}

\begin{equation*}
    \begin{split}
    \hat{f}_{s}^{l} & = \textrm{EMSA}(\textrm{LN}(f_{s}^{l-1})) + f_{s}^{l-1}\\
     f_{s}^{l} & = \textrm{MLP}(\textrm{GELU}(\textrm{Conv}_{3 \times 3}(\textrm{MLP}(\hat{f}_{s}^{l})))) + \hat{f}_{s}^{l}\\
     \hat{f}_{t}^{l} & = \textrm{EMSA}(\textrm{LN}(f_{t}^{l-1})) + f_{t}^{l-1}\\
     f_{t}^{l} & = \textrm{MLP}(\textrm{GELU}(\textrm{Conv}_{3 \times 3}(\textrm{MLP}(\hat{f}_{t}^{l})))) + \hat{f}_{t}^{l}\\
     \hat{f}_{t-s}^{l} & = \textrm{EMCA}(\textrm{LN}(f_{t}^{l-1}), \textrm{LN}(f_{s}^{l-1})) + f_{t-s}^{l-1}\\
     f_{t-s}^{l} & = \textrm{MLP}(\textrm{GELU}(\textrm{Conv}_{3 \times 3}(\textrm{MLP}(\hat{f}_{t-s}^{l})))) + \hat{f}_{t-s}^{l}\\
     \hat{f}_{s-t}^{l} & = \textrm{EMCA}(\textrm{LN}(f_{s}^{l-1}), \textrm{LN}(f_{t}^{l-1})) + f_{s-t}^{l-1}\\
     f_{s-t}^{l} & = \textrm{MLP}(\textrm{GELU}(\textrm{Conv}_{3 \times 3}(\textrm{MLP}(\hat{f}_{s-t}^{l})))) + \hat{f}_{s-t}^{l}\\
     \end{split}
\end{equation*}

\noindent where, $f_{s}^{l-1}$, $f_{s}^{l-1}$, $f_{t-s}^{l-1}$ and $f_{s-t}^{l-1}$ are the inputs for the $l$th quadruple transformer block, 'EMSE' and 'EMCA' denote the efficient multi-head self-attention and efficient multi-head cross attention, respectively.

\subsection{Cross-Domain Decoder}\label{cross_domain_decoder}
The cross-domain decoder generates in-domain and cross-domain features. Multi-scale features from each branch of the quadruple transformer encoder are passed through an MLP layer to make sure the channel dimensions are the same and are up-sampled to unify the resolution. Then, we utilize augmented feature representations $[\phi_{s}, \phi_{t-s}]$ and $[\phi_{t}, \phi_{s-t}]$ to predict segmentation masks. $\phi_{s}, \phi_{t-s}, \phi_{s-t}, \phi_{t}$ are of dimensions $[{H}/{4}, {W}/{4}, 4C]$ and the augmented features are of dimension $[{H}/{4}, {W}/{4}, 8C]$, which are then passed through another MLP layer to fuse the features. Finally, they are passed through the fourth MLP layer to predict segmentation masks, $\textrm{M}_{s}$ and $\textrm{M}_{t}$ for the source and target domain, respectively.

\subsection{Pseudo Label Correction}
One of the challenges in self-training is the generation of accurate pseudo labels for the target dataset. The generated pseudo labels are noisy, hence making the network fail to learn reliable information in the target domain. To address this issue, we implement a pseudo label correction mechanism. We draw inspiration from the ProDA \cite{zhang2021prototypical} to correct the noisy pseudo labels on the fly. Instead of computing prototypes from features of the target domain, we propose to generate cross-attentive prototypes that are computed from the features of the augmented representation. A cross-attentive prototype for a class $c$ is computed as a weighted sum of features from the augmented representation for the target domain and is given by - 

\begin{gather}
    \hat{f}_{c} = \dfrac{1}{\sum_{f \in F^{c}_{aug-t}} w(f, c)} \sum_{f \in F_{aug-t}^{c}} w(f, c) \cdot f
\end{gather}

where $F^{c}_{aug-t}$ are features from the augmented representation for the target domain, and $w(f, c)$ is the softmax probability of the respective pixel as provided by the pseudo labels. Since calculating prototypes this way are computationally expensive, we estimate them as a moving average of the semantic cluster centroids in mini-batches.

\begin{gather}
    \eta^{c} \longleftarrow \lambda \eta^{c} + (1 - \lambda)\eta^{'(c)}
\end{gather}

\noindent where $\eta^{'(c)}$ is the computed mean of features of class $k$ within the current training batch and $\lambda$ is the momentum coefficient which is set to 0.9999.

\mycomment{
Additionally, to avoid trivial solutions, the soft labels are fixed and are weighted by class-wise probabilities on the fly. The pseudo label correction scheme is formulated as - 

\begin{gather}
    \hat{y}_{t}^{c} = \xi (k_{t}^{k}p_{t, 0}^{c})
\end{gather}

where, $k_{t}^{k}$ is the weighting scheme that is given by the softmax over feature distances to prototypes. $p_{t, 0}^{c}$ is initialized by the warm-up model on the target images and remains fixed throughout the uda training stage. $\xi$ is the conversion from the soft predictions to hard labels by choosing only those pixels whose prediction confidence is higher than a given threshold. 
}

\subsection{Two-Way Image Pairing}
To ensure maximum context transfer between the two data distributions, we propose a two-way image pairing strategy for the cross-attention module. For each image in the source domain, we identify the most similar image from the target domain. The set of selected pairs would be given by - 

\begin{gather}
    \mathbb{P}_{s} = \{(x_{s}, x_{t})| x_{t} = \min_{x_{t}} d(x_{s}, x_{t}), \forall x_{s} \in X_{s}, \forall x_{t} \in X_{t}\}
\end{gather}

\noindent where $d(x_{s}, x_{t})$ is the structural similarity \cite{wang2004image} between image $x_{s}$ and $x_{t}$. Additionally, to eliminate the training bias from the target dataset, we augment more pairs $\mathbb{P}_{t}$ from the other way. Thus, for each image in the target domain, we identify the most similar image from the source domain which is given by -  

\begin{gather}
    \mathbb{P}_{t} = \{(x_{s}, x_{t})| x_{s} = \min_{x_{s}} d(x_{t}, x_{s}), \forall x_{s} \in X_{s}, \forall x_{t} \in X_{t}\}
\end{gather}

\noindent The final set of image pairs is obtained by taking the union of the two sets, i.e $\mathbb{P}=\{ \mathbb{P}_{s} \cup \mathbb{P}_{t} \}$.

\subsection{Training Objective}
The proposed methodology contains a segmentation loss $\mathcal{L}_{seg}$ and an adversarial loss $\mathcal{L}_{adv}$. The segmentation loss of $M_{s}$ is formulated as - 

\begin{gather}
    \mathcal{L}_{seg}(M_{s}, Y_{s}) = - \sum_{i=1}^{H\times W} \sum_{c=1}^{C} Y_{s}^{i, c} \log M_{s}^{i, c}
\end{gather}

\noindent where $C$ is the total number of semantic classes. Similarly, segmentation loss $\mathcal{L}_{seg}(M_{t}, Y_{t}^{st})$ is defined for the target segmentation mask. Furthermore, in order to adapt the structured output space \cite{tsai2018learning}, we utilize a discriminator to make the source and target masks indistinguishable from each other. To achieve this, we utilize an adversarial loss - 

\begin{gather}
    \mathcal{L}_{adv} (M_{s}, M_{t}, D) = \mathop{\mathbb{E}}[\log D(M_{s})] +  \mathop{\mathbb{E}}[\log 1 - D(M_{t})]
\end{gather}

\noindent Therefore, the total loss is formulated as - 

\begin{gather}
     \mathcal{L}_{total} = \mathcal{L}_{seg} (M_{s}, Y_{s}) + \beta_{1}*\mathcal{L}_{seg}((M_{t}, Y^{st}_{t}))  +  \beta_{2}*\mathcal{L}_{adv}
\end{gather}

\noindent The weighting coefficients $\beta_{1}$ and $\beta_{2}$ are set to
0.1 and 1, respectively.

\subsection{Inference for Target Domain}
During the inference process on the target distribution after the UDA stage, the inference scheme needs to use the source data. However, there is a storage cost to access the source data. Additionally, it is possible that the source data is not always available to us during inference. Hence, we propose an inference process that is independent of the source data. Since the QuadFormer cannot combine the augmented target feature representation (i.e, $[\phi_{t}, \phi_{s-t}]$) without the source data, we augment the target-aware features with itself, i.e $[\phi_{t}, \phi_{t}]$ to predict the segmentation mask $M_{t}$ during inference.

\begin{table}[h]
\caption{Comparison of state-of-the-art models on ARPLReal dataset.} 
\label{ARPLReal_seg_benchmark}
\centering 
\begin{tabular}{l c c c} 
\hline\hline 
 Method & Architecture & Params (M) & IoU
\\ [0.5ex]
\hline 
\raisebox{-1ex}{FCN} & \raisebox{-1ex}{ResNet-101} & \raisebox{-1ex}{68.8} & \raisebox{-1ex}{40.32}  \\[1ex]
\raisebox{-1ex}{UNet++} & \raisebox{-1ex}{Reset-101} & \raisebox{-1ex}{68.8} & \raisebox{-1ex}{49.87}  \\[1ex]

\raisebox{-1ex}{DeepLabv3+} & \raisebox{-1ex}{Reset-101} & \raisebox{-1ex}{68.8} & \raisebox{-1ex}{47.62}  \\[1ex]

\raisebox{-1ex}{PSPNet} & \raisebox{-1ex}{Reset-101} & \raisebox{-1ex}{68.8} & \raisebox{-1ex}{42.34}  \\[1ex]

\raisebox{-1ex}{Segformer} & \raisebox{-1ex}{MiT-B5} & \raisebox{-1ex}{84.7} & \raisebox{-1ex}{55.65}  \\[1ex]

\hline 
\end{tabular}
\label{tab:PPer}
\end{table}

\begin{figure*}[t]
\vspace{+2mm}
    \centering {\includegraphics[width=0.90\textwidth, trim={0 1cm 0 3cm},clip]{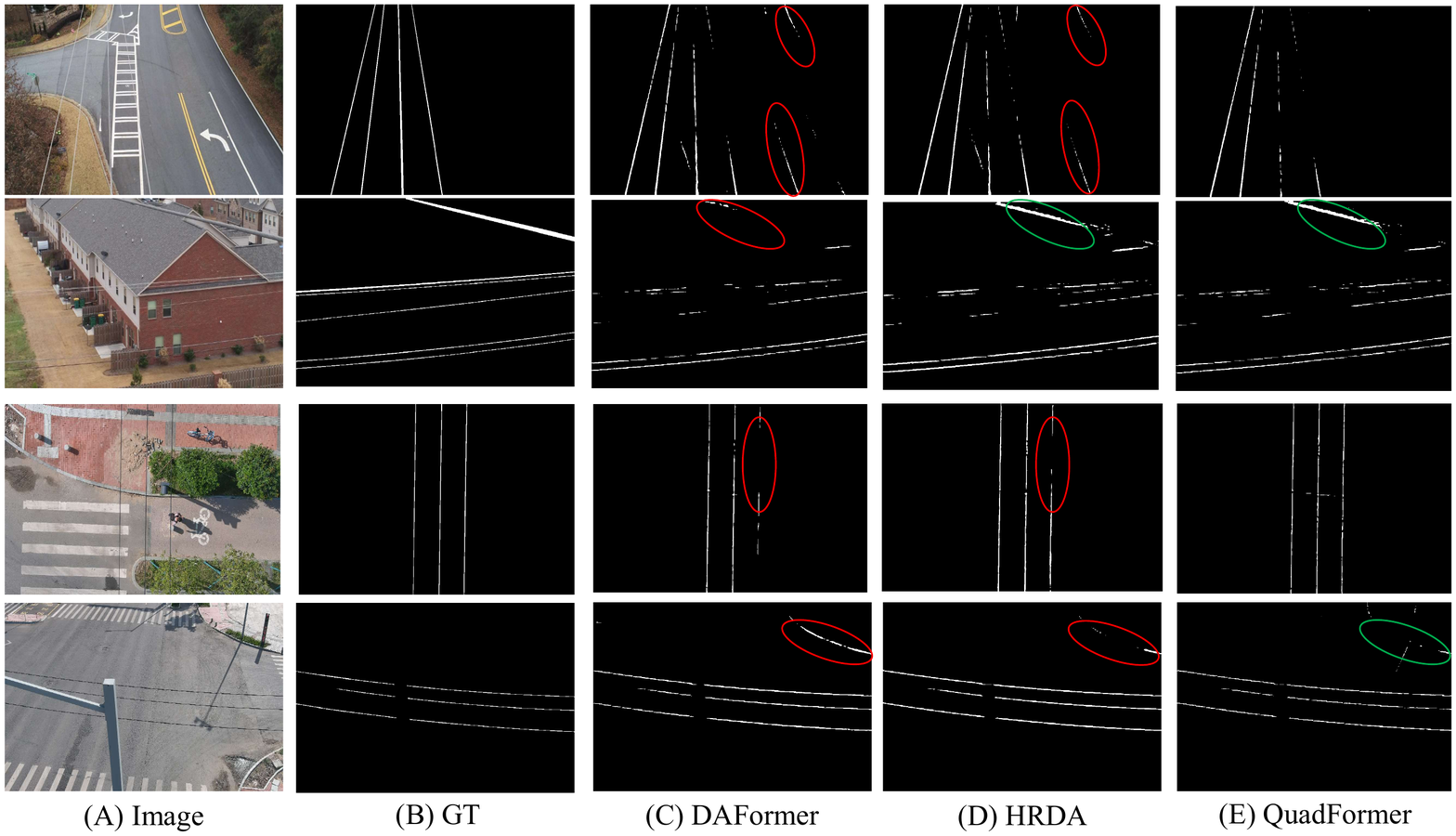}} \quad
    \vspace{-10mm}
    \caption{Qualitative analysis of validation images from TTPLA (first two rows) and ARPLReal (last two rows), when training models on the ARPLSyn datset. The red circle indicates incorrect predictions and the green circle indicated true positives.}
    \label{qualitative_analysis}
\vspace{-2mm}
\end{figure*}

\section{Experiments}
In this section, we evaluate the QuadFormer on synthetic-to-real domain adaptation for the power line segmentation task. We perform extensive experiments and ablation studies to demonstrate the performance of our model and compare its performance against existing state-of-the-art models.

\subsection{Implementation Details}
\noindent \textbf{Datasets} For the target domain, we utilize either the ARPLReal dataset containing 3800 training and 200 validation images with resolution $3840 \times 2160$, or the TTPLA dataset \cite{abdelfattah2020ttpla} which contains 1004 training images and 217 testing images with resolution $3840 \times 2160$. For the source domain, we use the ARPLSyn dataset, which contains 7000 synthetic images of resolution $3840 \times 2160$. As a common practice in UDA, we resize the images to $1024 \times 512$.

\noindent \textbf{Network Architecture} Our method is based on the mmsegmentation framework \cite{mmseg2020}. We use the MiT-B0 and MiT-B5 pre-trained on ImageNet-1k as the backbone network for the QuadFormer framework by following a similar setting in \cite{li2019bidirectional} as a warm-up. The decoder uses $C_{e} = 256$. The discriminator used for adapting the structured output has 5 convolutional layers with kernel $4 \times 4$ and stride of 2. Each layer has channels ${64, 128, 256, 512, 1}$.  

\noindent \textbf{Training} Similar to \cite{liu2021swin, xie2021segformer}, the QuadFormer is trained with AdamW \cite{loshchilov2017decoupled}, a learning rate of $\eta_{base} = 6 \times 10^{-5}$, a weight decay of 0.01, linear learning rate warmup ($t_{warm} = 1.5K$), followed by a linear decay. During training, data augmentation was applied through random horizontal flipping, photometric distortion, and random cropping to $512 \times 512$. The model was trained for 80K iterations. $\lambda$ is set to 0.1. All experiments were conducted on 4 NVIDIA GeForce RTX 3090 with PyTorch implementation.

\begin{table}[h]
\caption{Comparision results of ARPLSyn$\rightarrow$TTPLA adaptation in terms of IoU.} 
\label{ARPLSyn2TTPLA}
\centering 
\begin{tabular}{l c c c} 
 \hline
 \multicolumn{4}{c}{ARPLSyn$\rightarrow$TTPLA}
\\ [0.5ex]
\hline\hline 
 Method & Architecture & Params (M) & IoU
\\ [0.5ex]

\hline 
\raisebox{-1ex}{DACS \cite{tranheden2021dacs}} & \raisebox{-1ex}{ResNet-101} & \raisebox{-1ex}{68.8} & \raisebox{-1ex}{32.65}  \\[1ex]
\raisebox{-1ex}{DAFormer \cite{hoyer2022daformer}} & \raisebox{-1ex}{MiT-B5} & \raisebox{-1ex}{84.7} & \raisebox{-1ex}{40.21}  \\[1ex]

\raisebox{-1ex}{HRDA \cite{hoyer2022hrda}} & \raisebox{-1ex}{MiT-B5} & \raisebox{-1ex}{84.7} & \raisebox{-1ex}{43.55}  \\[1ex]

 \hline
\raisebox{-1ex}{Source-only} & \raisebox{-1ex}{MiT-B5} & \raisebox{-1ex}{84.7} & \raisebox{-1ex}{34.5}  \\[1ex]

\raisebox{-1ex}{\textbf{Ours}} & \raisebox{-1ex}{\textbf{MiT-B5}} & \raisebox{-1ex}{\textbf{84.7}} & \raisebox{-1ex}{\textbf{46.32}}  \\[1ex]

\hline 
\end{tabular}
\label{tab:PPer}
\end{table}

\subsection{Performance Comparison}
First, we benchmark state-of-the-art semantic segmentation methods on the ARPLReal dataset in Table.~\ref{ARPLReal_seg_benchmark}. Additionally, we evaluate the proposed QuadFormer on synthetic-to-real domain adaptation - a) ARPLSyn$\rightarrow$TTPLA and b) ARPLSyn$\rightarrow$ARPLReal. Furthermore, we conduct extensive experiments and perform ablation studies to demonstrate the superiority of our model over several recent leading UDA methods. Lastly, we provide qualitative results for ARPLSyn$\rightarrow$TTPLA and ARPLSyn$\rightarrow$ARPLReal. Figure.~\ref{qualitative_analysis} illustrates predictions of a few frames from the TTPLA and ARPLReal validation dataset made by QuadFormer-B5 and other models.

\noindent \textbf{Semantic Segmentation} Table.~\ref{ARPLReal_seg_benchmark} summarizes the performance of various state-of-the-art semantic segmentation methods on the ARPLReal dataset. In this experiment, we utilize ground truth annotations for bench-marking their performance. As shown, Segformer-B5 yields 55.65 IoU, outperforming all other baselines. For instance, compared to UNet++ and DeepLabV3, Segformer-B5 shows an improvement of 5.87 IoU and 7.7 IoU, respectively. We identify that compared to Segformer, which utilizes a hierarchically structured Transformer encoder for semantic segmentation, other models produce more false positives. This is attributed to the fact that the PL segmentation task is highly susceptible to being fragmented, and global context is essential for accurate predictions.

\noindent \textbf{ARPLSyn$\rightarrow$TTPLA}  
We first evaluate our method by utilizing ARPLSyn as the source domain and TTPLA as the target domain. The performance is assessed based on the model's ability to predict pixels corresponding to the PL class on the TTPLA validation set. We adopt the IoU metric to evaluate the model performance. Our method is compared with existing state-of-the-art models by using MiT-B5 as the backbone architecture. As indicated in Table.~\ref{ARPLSyn2TTPLA}, the QuadFormer achieves a state-of-the-art performance of 46.32 IoU, outperforming other baselines. Specifically, we surpass the IoU of DACS \cite{tranheden2021dacs} by 15.67 IoU. Additionally, it is observed that the transformer-based techniques perform better than DACS because they leverage global context. Since the PL segmentation task involves datasets which are highly unbalanced, global information is necessary. Compared to transformer-based UDA models \cite{hoyer2022daformer, hoyer2022hrda}, our method gains up to 4.58 IoU improvement by utilizing cross-attentive features, revealing that domain discrepancy can be further reduced by considering context adaptation.

\begin{figure*}[t]
\vspace{+2mm}
    \centering {\includegraphics[width=0.90\textwidth, trim={0 1.20cm 0 3cm},clip]{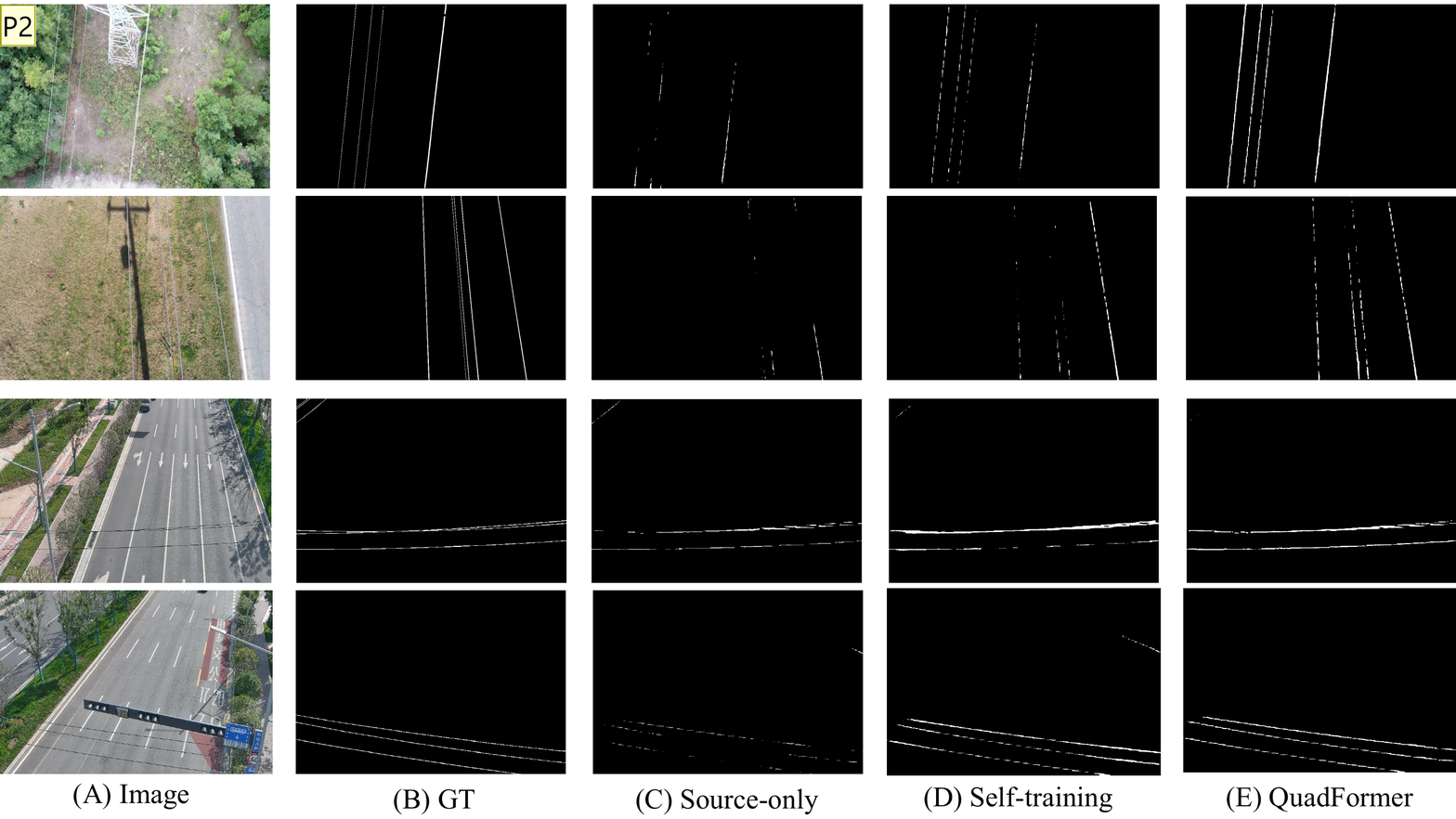}}
    \quad
    \vspace{-10mm}
    \caption{Qualitative analysis of key components of the proposed QuadFormer on validation images from TTPLA (first two rows) and ARPLReal (last two rows). Predictions from the source-only model (C) illustrate the domain gap during knowledge transfer. After enabling self-training with cross-domain features (D), the predictions are more accurate. Finally, the proposed pseudo-label correction mechanism (E) online denoises the pseudo labels and refines the prediction.}
    \label{illustration_ablation}
\end{figure*}

\begin{table}[h]
\caption{Comparision results of ARPLSyn$\rightarrow$ARPLReal adaptation in terms of IoU.} 
\label{ARPLSyn2ARPLReal}
\centering 
\begin{tabular}{l c c c} 
 \hline
 \multicolumn{4}{c}{ARPLSyn$\rightarrow$ARPLReal}
\\ [0.5ex]
\hline\hline 
 Method & Architecture & Params (M) & IoU
\\ [0.5ex]

\hline 
\raisebox{-1ex}{DACS \cite{tranheden2021dacs}} & \raisebox{-1ex}{ResNet-101} & \raisebox{-1ex}{68.8} & \raisebox{-1ex}{30.27}  \\[1ex]
\raisebox{-1ex}{DAFormer \cite{hoyer2022daformer}} & \raisebox{-1ex}{MiT-B5} & \raisebox{-1ex}{84.7} & \raisebox{-1ex}{39.67}  \\[1ex]

\raisebox{-1ex}{HRDA \cite{hoyer2022daformer}} & \raisebox{-1ex}{MiT-B5} & \raisebox{-1ex}{84.7} & \raisebox{-1ex}{42.85}  \\[1ex]

 \hline
\raisebox{-1ex}{Source-only} & \raisebox{-1ex}{MiT-B5} & \raisebox{-1ex}{84.7} & \raisebox{-1ex}{35.90}  \\[1ex]

\raisebox{-1ex}{\textbf{Ours}} & \raisebox{-1ex}{\textbf{MiT-B5}} & \raisebox{-1ex}{\textbf{84.7}} & \raisebox{-1ex}{\textbf{44.25}}  \\[1ex]

\hline 
\end{tabular}
\label{tab:PPer}
\end{table}

\noindent \textbf{ARPLSyn$\rightarrow$ARPLReal} The domain gap between ARPLSyn and ARPLReal is greater than that of ARPLSyn and TTPLA due to the highly diverse nature of the ARPLReal dataset. In Table.~\ref{ARPLSyn2ARPLReal}, we show the adaptation results on the ARPLReal validation set, where QuadFormer shows considerable improvement. The proposed QuadFormer with the MiT-B5 backbone achieves an IoU of 44.25 and outperforms all baselines. This further demonstrates the benefit of leveraging cross-domain context in semantic segmentation with highly unbalanced datasets.

\begin{table}[h]
\caption{Ablation study on the ARPLSync$\rightarrow$ARPLReal adaptation to understand the contribution of context adaptation. We use QuadFormer-B0 and train it for 40K iterations.} 
\centering 
\label{ablation_feat}
\begin{tabular}{l c c c} 
 \hline
 
 Source features & Target features  & IoU
\\ [0.5ex]

\hline 
\raisebox{-1ex}{self-attention} & \raisebox{-1ex}{self-attention} & \raisebox{-1ex}{34.62}  \\[1ex]
\raisebox{-1ex}{cross-attention} & \raisebox{-1ex}{self-attention} & \raisebox{-1ex}{36.34} \\[1ex]

\raisebox{-1ex}{self-attention} & \raisebox{-1ex}{cross-attention} & \raisebox{-1ex}{37.48} \\[1ex]

\raisebox{-1ex}{cross-attention} & \raisebox{-1ex}{cross-attention} & \raisebox{-1ex}{39.65}  \\[1ex]

\hline 
\end{tabular}
\label{tab:PPer}
\end{table}

\subsection{Ablation study}
We perform extensive ablation studies to demonstrate the key components of our proposed UDA model. In all ablation studies, we use our lightweight model, QuadFormer-B0 on ARPLSync$\rightarrow$ARPLReal and train the model for 40K iterations.

\noindent \textbf{Effect of cross-attention} In Table.~\ref{ablation_feat}, we perform ablation studies to study the effect of source and target features using only the self-attention and using both the self-attention and cross-attention. In contrast to only self-attentive features in both domains, incorporating cross-attention in the source features or target features improves the IoU by 1.72 $\%$ and 2.86 $\%$, respectively. By introducing cross-attention in both domains, we achieve 39.65 IoU. While the self-attentive prototypes are able to correct the noisy pseudo labels up to some extent, the cross-attentive prototypes are less sensitive to outliers due to context adaptation. Hence, the cross-domain context at the feature level is important to understand the semantic distribution in both domains and reduces the discrepancy in data distributions.

\begin{table}[h]
\caption{Ablation study of each proposed component on the ARPLSync$\rightarrow$ARPLReal adaptation. We use QuadFormer-B0 and train it for 40K iterations.} 
\label{ablation_component}
\centering 
\begin{tabular}{l c c c c} 
 \hline
 
self & adversarial  & pseudo label & IoU
\\ [0.5ex]
training & loss  & denoising & 
\\ [0.5ex]

\hline 
\raisebox{-1ex}{} & \raisebox{-1ex}{} & \raisebox{-1ex}{} & \raisebox{-1ex}{25.67} \\[1ex]
\raisebox{-1ex}{\checkmark} & \raisebox{-1ex}{} & \raisebox{-1ex}{} & \raisebox{-1ex}{33.5} \\[1ex]
\raisebox{-1ex}{\checkmark} & \raisebox{-1ex}{\checkmark} & \raisebox{-1ex}{} & \raisebox{-1ex}{35.56}\\[1ex]

\raisebox{-1ex}{\checkmark} & \raisebox{-1ex}{\checkmark} & \raisebox{-1ex}{\checkmark} & \raisebox{-1ex}{39.65}\\[1ex]

\hline 
\end{tabular}
\label{tab:PPer}
\end{table}

\noindent \textbf{Effect of key components of the UDA framework} Table.~\ref{ablation_component} indicates an ablation study of each proposed component. The source-only (SegFormer-B0) gives 25.67 IoU on the target domain. Initialized by the source-only model, self-training with QuadFormer-B0, where the model is trained on offline pseudo-labels achieves an IoU gain of 7.83. Adding the adversarial loss brings a 2.06 IoU gain. Finally, with cross-attentive prototypes for online pseudo-label correction and all other components, our model achieves 39.65 IoU. The cross-attentive prototype correction scheme treats all semantic classes equally. This significantly improves the performance of PL segmentation. Figure.~\ref{illustration_ablation} illustrates the effect of key components of the proposed QuadFormer.

\section{Conclusion}
In this work, we propose a novel UDA framework, QuadFormer, for cross-domain PL segmentation. Specifically, the proposed framework leverages cross-domain context dependencies to online denoise the pseudo labels. The quadruple transformer integrates cross-attention and self-attention at multiple levels for learning domain-invariant feature representations. Additionally, we present two datasets, ARPLSyn and ARPLReal to promote research in domain adaptive PL segmentation. Lastly, extensive experiments indicate that our best performing model, QuadFormer-B5 achieves state-of-the-art performance for the domain adaptive power line segmentation on ARPLSyn$\rightarrow$TTPLA and ARPLSyn$\rightarrow$ARPLReal.

\clearpage
{\small
\bibliographystyle{ieee_fullname}
\bibliography{egbib}
}

\end{document}